\documentclass[12pt]{article}
\usepackage{tikz}
\usepackage{arxiv}

\usepackage[american]{babel} 

\usepackage[utf8]{inputenc} 
\usepackage[T1]{fontenc} 
\usepackage{lmodern}
\usepackage{hyperref}       
\usepackage{url}            
\usepackage{booktabs}       
\usepackage{amsfonts}       
\usepackage{nicefrac}       
\usepackage{microtype}      
\usepackage[numbers]{natbib}
\usepackage{doi}
\usepackage{siunitx}
\usepackage{amsmath}
\usepackage{amssymb}
\usepackage{cleveref}       
\usepackage{xcolor,graphicx,eso-pic}
\usepackage{lipsum}
\usepackage{arydshln}
\usepackage{bera} 
\usepackage[export]{adjustbox}

\hypersetup{colorlinks}

\newcommand\ourPaperTitle{Speaking the Same Language: Leveraging LLMs in Standardizing Clinical Data for AI}

\newcommand\ourKeywords{Large Language Models \and Data Engineering \and Common Data Model (CDM) \and{HL7 FHIR}\and Data Standardization \and Clinical Research \and Artificial Intelligence}

\title{\ourPaperTitle}

\author{
    Arindam Sett\thanks{Corresponding Author} \\
    \texttt{sett.arindam@gene.com} \\
    \And 
    Somaye Hashemifar \\
    \texttt{hashemifar.somaye@gene.com} \\
    \And
    Mrunal Yadav \\
    \texttt{yadavm14@roche.com} \\
    \And 
    Yogesh Pandit\thanks{Co-Principal Investigator. For Mr. Hejrati, work done while employed at Genentech.} \footnotemark[1]  \\
    \texttt{pandit.yogesh@gene.com} \\
    \And
    Mohsen Hejrati\footnotemark[2] \\
    \texttt{mhejrati@gmail.com} \\
    \And 
    \\
    \text{gRED Computational Sciences (gCS)}\\
    \text{Genentech Research and Early Development (gRED)}\\
    \text{Genentech} \\
    \text{South San Francisco, CA, USA} \\
    \\
    \textbf{for the Swedish BioFINDER study group}\thanks{A complete list of the BioFINDER study group members can be found at \url{http://www.biofinder.se}.} \\
    \AND 
    for the Alzheimer's Disease Neuroimaging Initiative\thanks{Data used in preparation of this article were obtained from the Alzheimer's Disease Neuroimaging Initiative (ADNI) database (\href{http://adni.loni.usc.edu}{adni.loni.usc.edu}). As such, the investigators within the ADNI contributed to the design and implementation of ADNI and/or provided data but did not participate in the analysis or writing of this report. A complete listing of ADNI investigators can be found at \url{http://adni.loni.usc.edu/wpcontent/uploads/how_to_apply/ADNI_Acknowledgement_List.pdf} } \\
}

\begin{document}

\maketitle
\begin{abstract}
        The implementation of Artificial Intelligence (AI) in the healthcare industry has garnered considerable attention, attributable to its prospective enhancement of clinical outcomes, expansion of access to superior healthcare, cost reduction, and elevation of patient satisfaction. Nevertheless, the primary hurdle that persists is related to the quality of accessible multi-modal healthcare data in conjunction with the evolution of AI methodologies. This study delves into the adoption of large language models to address specific challenges, specifically, the standardization of healthcare data. We advocate the use of these models to identify and map clinical data schemas to established data standard attributes, such as the Fast Healthcare Interoperability Resources. Our results illustrate that employing large language models significantly diminishes the necessity for manual data curation and elevates the efficacy of the data standardization process. Consequently, the proposed methodology has the propensity to expedite the integration of AI in healthcare, ameliorate the quality of patient care, whilst minimizing the time and financial resources necessary for the preparation of data for AI.
\end{abstract}


    \section{Introduction}\label{sec:introduction}

        Using clinical datasets from disparate sources with different schema (data structures) and data dictionaries (information about fields, data types, and their meanings) can pose several challenges in AI development:
        \begin{itemize}
            \item \textbf{Data Inconsistency}

            Different sources may have different conventions for documenting the same information. This inconsistency can lead to data mishandling, which in turn jeopardizes the accuracy of AI models.

            \item \textbf{Feature Mismatch}

            If different datasets use different features, or represent identical features differently, it can be difficult to reconcile them into a uniform model. Even similar types of data can be represented differently, making harmonization challenging.

            \item \textbf{Data Quality}
            
            Datasets from different sources often have different levels of quality and reliability. Some datasets may contain more errors or inaccuracies than others, which can affect the performance of the AI models.
            
            \item \textbf{Information Loss}
            
            During the process of standardizing or transforming the data to fit into a single schema, some important information could be lost.
            
            \item \textbf{Higher Complexity}
            
            Managing data from disparate sources increases the overall complexity of data pre-processing. This also means increased time and resources spent on data wrangling.
            
            \item \textbf{Legal and Ethical Issues}
            
            The use of disparate data sources raises concerns about privacy and data security. Furthermore, the algorithms developed using the data can be unknowingly biased due to the variations in the quality of data from these different sources.
            
            \item \textbf{Interoperability Issues}
            
            Different healthcare information systems tend to use different data standards, creating compatibility issues and limiting how freely data can be exchanged between systems.
        \end{itemize}

        Ultimately, such challenges may compromise the validity, generalizability, and reliability of AI solutions in the clinical field. This emphasizes the importance of a robust, harmonized data sourcing and integration strategy in clinical AI development.

        Previous research has extensively explored the use of AI in clinical data standardization, primarily focusing on rule-based methodologies and conventional machine learning techniques \citep{modaresnezhad2019, papadopoulos2022}. These approaches have paved the way for notable advancements in handling structured data. However, they often fall short when dealing with complicated domain-specific schemas, which constitute a significant portion of real-world clinical datasets. Techniques like natural language processing have been applied to extract information from unstructured text \citep{sezgin2023, adamson2023} or data cleansing \citep{cardinal2017}. Yet, the application of Large Language Models (LLM) has been relatively under-explored in the current literature. The potential of these models for semantic understanding and context-aware data mapping suggests they could offer substantial improvements in the field of clinical data standardization.

        This work is an attempt to leverage LLMs for mapping any raw dataset to a clinical data standard; HL7 Fast Healthcare Interoperability Resources (FHIR) \citep{hl7_fhir}. We leverage a zero/few-shot learning approach to achieve data mapping for standardization.

        \textbf{Note:} In course of this work, we have used only the clinical data column names and corresponding data dictionaries (which defines the columns, their data types, applicable code values etc.) where available, we have refrained from using actual clinical data as that is not a requirement for the experiments. Also, in some cases we have used input datasets curated in CDISC \citep{CDISC_Standards} Clinical Data standard rather than the original data models. However, our methods are agnostic of source data model.

\subsection{A Brief Overview of Data Standardization for Clinical Data}\label{subsec:data-standard-overview}

        The adoption of data standards for clinical data has become increasingly important with the rise of AI applications in healthcare. To develop accurate and effective AI models for tasks like diagnosis, prognosis, and treatment recommendations, high-quality structured clinical data is required. However, clinical data has historically been challenging to standardize due to its complexity and heterogeneity.

        In recent years, FHIR has emerged as a leading standard for clinical data exchange. FHIR provides a common framework and set of APIs for representing and sharing clinical data in a standardized way. Some key benefits of FHIR include:
        \begin{itemize}
            \item Structured data format based on resources with common fields for clinical concepts like patients, conditions, medications, etc. This enables integration and analysis across datasets. 
            \item Modular components can be used in flexible ways to represent various clinical workflows. This facilitates interoperability across systems. 
            \item Modern web standards and APIs for efficient data access and exchange.
            \item Active open source community with rapid evolution of specifications.
        \end{itemize}

\begin{figure}[h]
\begin{center}
\includegraphics[width=1\linewidth]{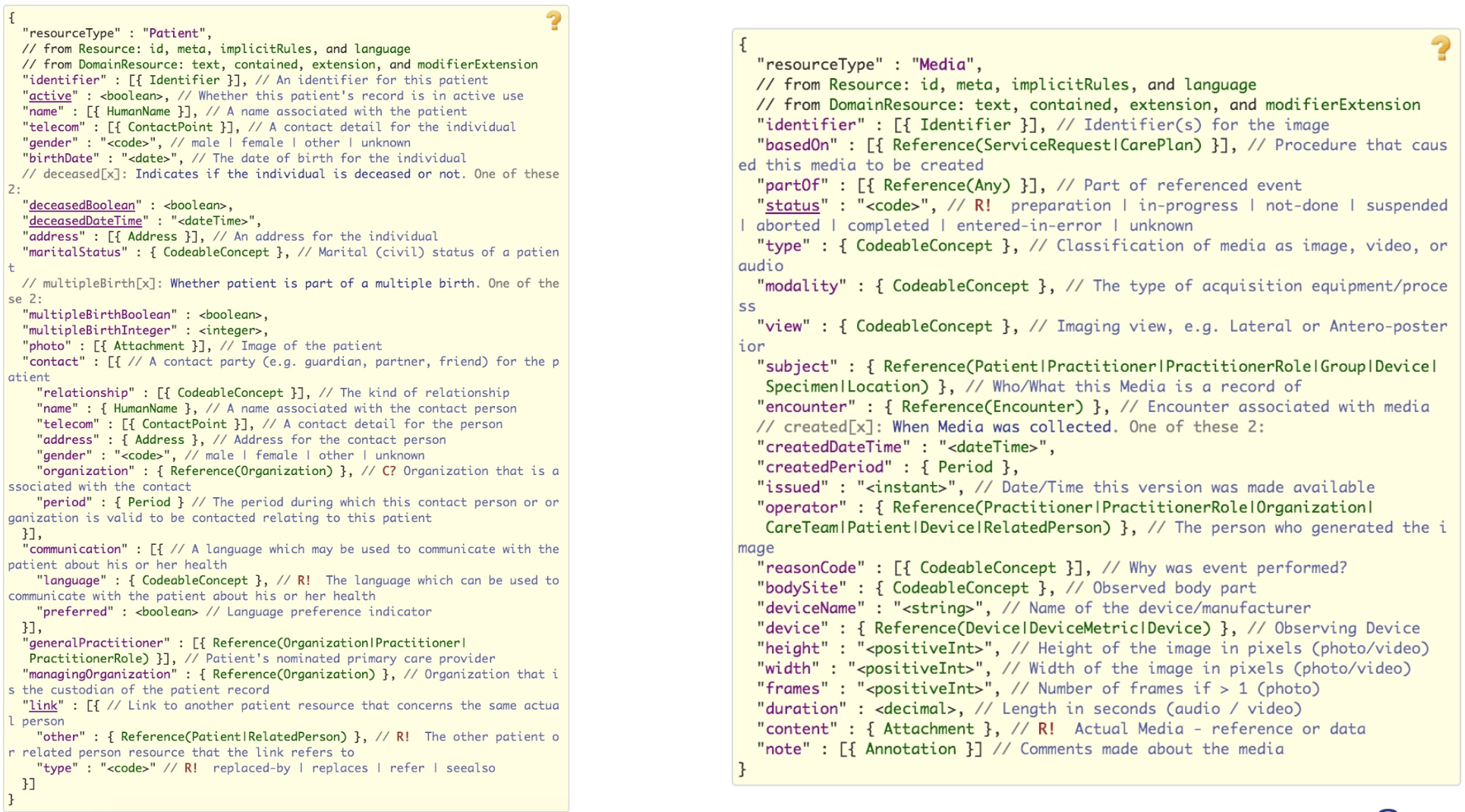}
\end{center}
\caption{An example of \texttt{Patient} and \texttt{Media} FHIR resources' schemas}
\label{fig:fhir_schema_example}
\end{figure}

       FHIR acts as a bridge for AI, allowing it to extract structured clinical data from electronic health records and other healthcare systems in a consistent format. This helps address the "data wrangling" challenges that often dominate healthcare AI projects. With more adoption of FHIR, higher-quality datasets will become available for developing, evaluating, and deploying AI algorithms. An example of FHIR representation of patient data and media data is shown in Figure~\ref{fig:fhir_schema_example} ~\citep{hl7_fhir}.

In summary, data standards like FHIR are foundational for realizing the potential of AI in healthcare. By enabling interoperability and data accessibility, FHIR can help guide the responsible and effective application of AI to drive better clinical outcomes and patient care. Ongoing adoption of FHIR will be crucial for advancing clinical AI innovation in the years ahead.
        
\section{Tasks and Data}\label{sec:methods}

\subsection{Data}\label{subsec:dataset}
This study utilized $14$ clinical datasets (details on access requirements for each dataset can be provided upon request). The datasets encompassed a wide range of disease areas, including neurology, respiratory, ophthalmology, and insurance claims. The size and complexity of the datasets varied considerably. Table~\ref{tab:dataset_table} shows Datasets and Number of Fields:

\begin{table}[h]
\begin{tabular}{|c|c|c|}
\hline
\textbf{Dataset} & \textbf{\# Fields} & \textbf{Therapeutic Area} \\
\hline
ADNI~\citep{petersen2010alzheimer} & 258 & \\
A4 ~\citep{a4study} & 514 & \\
AIBL ~\citep{ellis2009galantamine} & 101 & \\
BLAZE (NCT01397578)& 191 & Neurology \\
ADNI\_DOD ~\citep{adnidod} & 3254 & \\
BIOFINDER1~\citep{mattsson2020implications} & 1698 & \\
TAURIEL(NCT03289143) ~\citep{teng2022safety}   & 1398 & \\
HABS~\citep{dagley2017harvard} & 1093 & \\
\hdashline
NLST (NCT00047385) & 94 & \\
LUNG\_PET\_CT\_DX ~\citep{li2020large, clark2013cancer},  & 25 & Respiratory \\
COVASTIL (NCT04386616) & 902 & \\
LIDC-IDRI ~\citep{armato2015data,clark2013cancer, armato2011lung} & 25 & \\
\hdashline
CERA (NCT01790802) ~\citep{cera} & 51 & Ophthalmology \\
\hdashline
CITELINE ~\citep{CitelineRWD} & 138 & Claims Data \\
\hline
\end{tabular}
\caption{Summary of datasets}
\label{tab:dataset_table}
\end{table}

\subsubsection{Ground-truth Generation and Curation}
Ground-truth generation employed a semi-supervised approach leveraging LLMs. The resulting FHIR mapping was subsequently reviewed and refined by a team of four data curators with expertise in FHIR.

\subsection{Methods}\label{subsec:methods}
In this section, we describe the motivation behind our system, its overview and functionalities, the prompt engineering process.

\subsubsection{System Inspiration} 
Our goal is to explore the ability of GPT-3.5 to provide a framework for defining a consistent data terminology across various datasets and institutions by utilizing Retrieval-Augmented Generation (RAG)~\citep{DBLP:journals/corr/abs-2005-11401}. This standardization is essential for interoperability, enabling healthcare organizations to exchange patient data seamlessly, researchers to aggregate and analyze data consistently, and developers to create applications that work with healthcare information reliably.  

\subsubsection{FHIR Standard and Clinical Data Mapping to FHIR} 
The core foundation of FHIR Standard is a set of modular components called "Resources". A Resource represents instance-level representation of any healthcare entity. All resources have a few elements (also called attributes) in common: 

\begin{itemize}
\item An identifier for the resource - typically a URL that defines where the resource is found
\item Common metadata
\item A human-readable summary
\item A set of defined data elements - a different set for each type of resource
\end{itemize}

Resource instances are represented as either XML, JSON or RDF and there are currently 157 different resource types defined in the FHIR specification (FHIR v5.0.0: R5) \citep{FHIRv5Resources}

Clinical Data is inherently multi-modal, however, all clinical data have a textual representation of the meaning of a particular data element, which is usually captured in a data dictionary. A typical structure of a data dictionary is shown in Table~\ref{tab:compact_table}. Here we see along with the field name, we also have a short description for the field, oftentimes this is accompanied by a list of coded values for the field. Together with the field name, these constitute the input for the data mapping exercise. Format of ground-truth as well as Expected Output is shown in Table~\ref{tab:fhir_mapping}.

\begin{table}[h]
\begin{tabular}{|l|l|l|}
\hline
\textbf{dataset\_name} & \textbf{field\_name} & \textbf{field\_description} \\ \hline
ADNI & MAGSTRENGTH & MRI Machine Magnetic Field Strength \\ \hline
ADNI & BRAINSTEM & brain-stem \\ \hline
ADNI & BRAINSTEM\_SIZE & brain-stem ROI size in mm\textsuperscript{3} \\ \hline
ADNI & CC\_ANTERIOR & cc-anterior \\ \hline
ADNI & CC\_ANTERIOR\_SIZE & cc-anterior ROI size in mm\textsuperscript{3} \\ \hline
ADNI & CC\_CENTRAL & cc-central \\ \hline
ADNI & CC\_CENTRAL\_SIZE & cc-central ROI size in mm\textsuperscript{3} \\ \hline
\end{tabular}
\caption{A data dictionary structure}
\label{tab:compact_table}
\end{table}

\begin{table}[h]
\begin{tabular}{|l|l|l|}
\hline
\textbf{dataset\_name} & \textbf{field\_name} & \textbf{fhir\_mapping} \\ \hline
ADNI & MAGSTRENGTH & ImagingStudy.series.extension.valueDecimal \\ \hline
ADNI & BRAINSTEM & Observation.valueQuantity.value \\ \hline
ADNI & BRAINSTEM\_SIZE & Observation.component.valueQuantity.value \\ \hline
ADNI & CC\_ANTERIOR & Observation.component.code.coding.code \\ \hline
ADNI & CC\_ANTERIOR\_SIZE & Observation.component.valueQuantity.value \\ \hline
ADNI & CC\_CENTRAL & Observation.component.code.coding.code \\ \hline
ADNI & CC\_CENTRAL\_SIZE & Observation.component.valueQuantity.value \\ \hline
\end{tabular}
\caption{Dataset to FHIR mapping}
\label{tab:fhir_mapping}
\end{table}

\subsubsection{System Overview} 
The system takes the data dictionary of a target dataset as input and generates an output mapping individual data elements that conforms to the FHIR standard, resulting in a standardized format for the dataset. This data dictionary typically includes definitions, descriptions, and explanations of terms, fields, and variables used in the dataset.

\begin{figure}
\begin{center}
\includegraphics[width=1\linewidth]{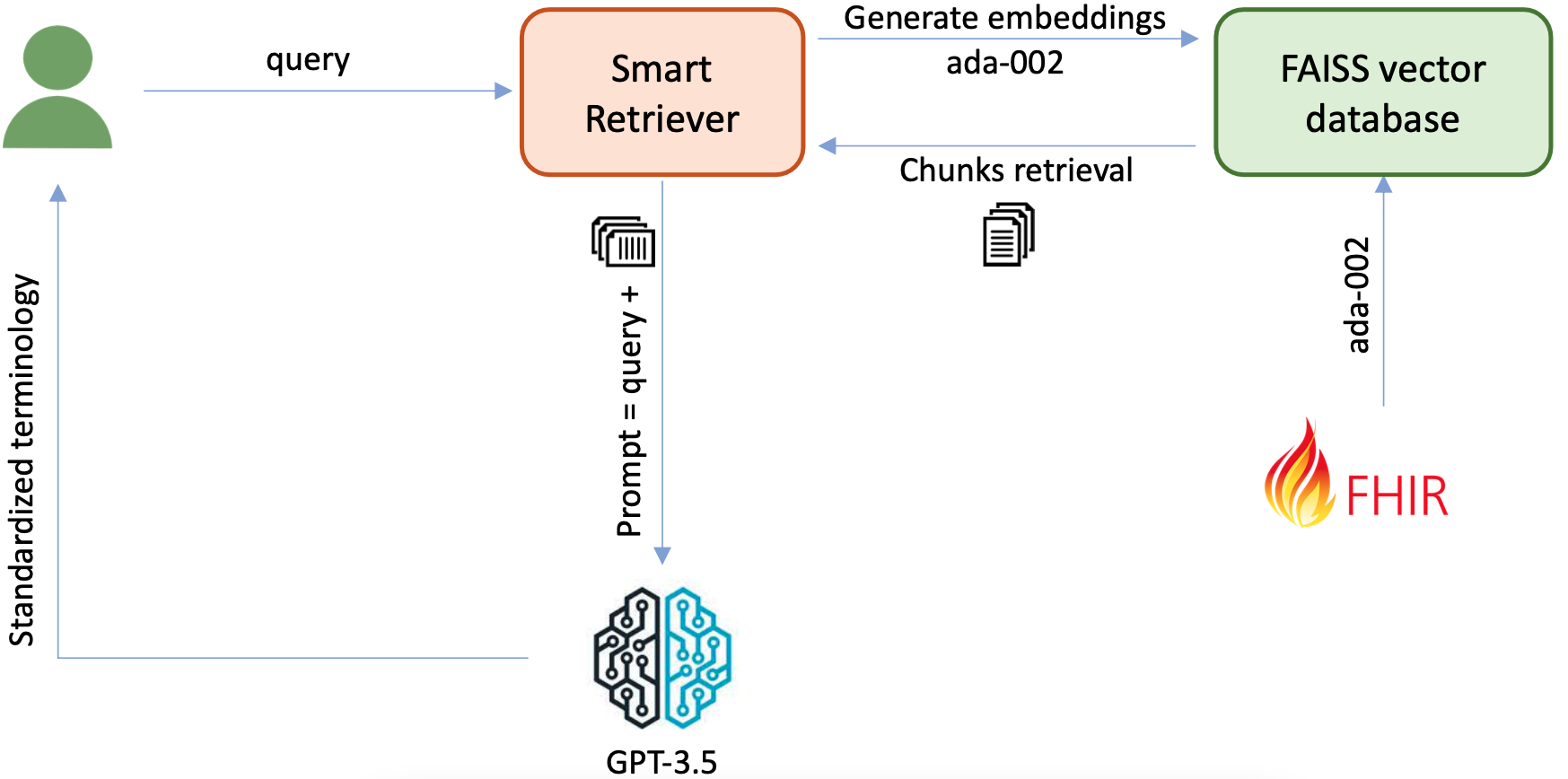}
\end{center}
\caption{An overview of our approach.}
\label{fig:overview}
\end{figure}

As shown in Figure~\ref{fig:overview}, our system combines a retriever system, which extracts relevant document snippets from FHIR, and GPT-3.5, which produces answers using the information from those snippets. In essence, RAG helps the model to search and retrieve contextual information from FHIR to improve its responses.
Specifically, the FHIR documents are split into text chunks using recursive character text splitting. The chunks are then embedded in a vector space using the OpenAI text-embedding-ada-002 embedding engine and stored in the Facebook AI Similarity Search (FAISS 1.7.4) vector database. The FAISS\citep{faiss} vector database is used to find the k-most similar chunks to a given query at the query time. The original query, combined with the retrieved chunks is compiled into a prompt and passed to the GPT-3.5 for generating the answer. This provides the LLM with additional information that contains factual data, which can help to improve the quality of its responses. For vector similarity search we have used a chunk size of 2000, chunk overlap of 200.

\subsubsection{Context for RAG}\label{subsec:RAG-context}
For RAG, we have prepared content based on the FHIR standard definitions that are available at the HL7 FHIR website. We have taken the resource, element and the corresponding descriptions of the resource-element pair as information that we have subsequently embedded into the vector database using chunking, for a subsequent retrieval during mapping, this retrieved context is passed along with the input data dictionary column and associated information to the LLM for an output FHIR mapping.

\subsubsection{Prompt Engineering}\label{subsec:prompt-engg}

Prompt engineering is an emerging field that focuses on creating and fine-tuning prompts to maximize the effectiveness of LLMs for various applications. It involves a comprehensive set of techniques that boost interaction with LLMs, enhance their safety, integrate domain knowledge, and work with external tools.

In this work we have taken an incremental approach with prompt engineering to create FHIR mapping. Initially we set out with basic prompting with minimal information in the prompt (viz. just the input data dictionary row and some simple instructions to map to FHIR), and then iterate over the input data dictionary sets after carefully examining output FHIR mapping generated. Our final prompt has the following structure: Definition of User Role, Initial Instructions, Placeholder for Retrieved Context from RAG, Example (one-shot learning example), Placeholder for data dictionary Input, Direction for Output Format, and Final Set of Instructions. Of these, we noticed the biggest change in output results occurred because of variations in Final Set of Instructions, as well content of the data dictionary.

\subsection{Preliminary Evaluation}\label{subsec:results}
Our model predicts the structures for the data dictionary of a dataset. Each structure consists of several metadata blocks, with the first block usually called the resource. We evaluated the performance of our model based on the number of matched metadata blocks between the predicted structure and the ground-truth. We evaluated the performance for top-$k$ = $20$.

A predicted structure is categorized as an 'Absolute Match' if all of its blocks align with those in the ground-truth. If the resource aligns but some or all other metadata elements do not, this leads to a 'Partial Match'. Alternatively, if none of the blocks match, or if the resource does not match even when other block do, the predicted structure is considered a 'Mismatch'. Partial Score and Match Score are then defined as below:

\begin{align*}
&Score = \frac{S + P}{N} \times 100 \\\\
P = \sum&_{i=1}^K \frac{\text{intersection}(pred_i, gt_i)}{\text{union}(pred_i, gt_i)}
\end{align*}

Where $S$ and $N$ represent the quantity of absolute matches and the total structures in the dataset dictionary, respectively. $P$ is the fraction of matched blocks in all 'Partial Match' structures. $pred_i$ and $gt_i$ correspond to the predicted and ground-truth for structure $i$, respectively. $K$ is the number of all 'Partial Match' structures. Moreover, we also calculate a Resource Match Score, which is calculated based on instances where the resources matched between ground-truth and Predicted results. Similar to Partial Score this is also a percentage value.

\begin{align*}
{Resource} \; {Match} \; {Score} = \frac{S + K}{N} \times 100 \\
\end{align*}

Where $S$, $K$, and $N$ represent the quantity of absolute matches, the quantity of 'Partial Match' structures, and the total structures in the dataset dictionary, respectively.

\begin{table}[ht]
\centering
\begin{tabular}{|c|c|c|}
\hline
\textbf{Dataset}      & \textbf{Score(\%)}                & \textbf{Resource Match Score(\%)} \\ \hline
A4                    & 57.16 ($\pm 0.77$)                & 94.36 ($\pm 0.34$) \\ \hline
ADNI                  & 51.31 ($\pm 1.01$)                & 66.58 ($\pm 1.24$) \\ \hline
ADNI\_DOD             & 77.93 ($\pm 0.2$)                 & 97.20 ($\pm 0.18$) \\ \hline
AIBL                  & 73.26 ($\pm 2.15$)                & 90.89 ($\pm 1.86$) \\ \hline
BIOFINDER1            & 77.09 ($\pm 0.53$)                & 95.84 ($\pm 0.45$) \\ \hline
BLAZE                 & 79.59 ($\pm 0.94$)                & 96.70 ($\pm 0.55$) \\ \hline
CERA                  & 73.17 ($\pm 3.05$)                & 89.8 ($\pm 2.75$) \\ \hline
CITELINE              & 64.18 ($\pm 2.9$)                 & 82.39 ($\pm 2.69$) \\ \hline
COVASTIL              & 78.05 ($\pm 0.36$)                & 96.03 ($\pm 0.38$) \\ \hline
HABS                  & 71.75 ($\pm 0.49$)                & 95.15 ($\pm 0.4$) \\ \hline
LIDC-IDRI             & 72.77 ($\pm 3.6$)                 & 84.00 ($\pm 3.77$) \\ \hline
NLST                  & 67.48 ($\pm 2.59$)                & 84.6 ($\pm 2.13$) \\ \hline
TAURIEL               & 71.20 ($\pm 0.47$)                & 94.09 ($\pm 0.28$) \\ \hline
LUNG\_PET\_CT\_DX     & 52.86 ($\pm 4.23$)                & 68.00 ($\pm 3.77$) \\ \hline
\textbf{Total}        & \textbf{73.54} ($\pm 0.16$)       & \textbf{94.52} ($\pm 0.11$) \\ \hline
\end{tabular}
\caption{Scores for Datasets Mapped to FHIR (top-k=20) averaged over 10 mapping iterations}
\label{tab:score_topk20}
\end{table}

\subsection{Observations}\label{subsec:observations}
Table~\ref{tab:score_topk20} presents the results for top-$k=20$, averaged over 10 mapping iterations to account for the variations in results in individual mapping run. 

The model exhibited strong overall performance with a mean score of $73.54$ ($SD = 0.16$) and a low standard deviation of $0.11$ for the resource match score ($94.52$). This consistency suggests that the model performs reliably across various datasets and exhibits minimal variability between runs.

The resource match score, which evaluates the alignment between predicted resources and the ground-truth, shows variability across datasets. While some datasets demonstrate high resource match scores (e.g., ADNI\_DOD with $97.20$), others exhibit lower alignment (e.g., CERA with $89.8$). The consistency in resource match scores, as indicated by small standard deviations, suggests that the model maintains a relatively stable performance in resource prediction across different datasets. 

some datasets exhibit notably high scores, such as ADNI\_DOD ($77.93$), COVASTIL ($78.05$), BIOFINDER1 ($77.09$), and BLAZE ($79.59$), while others, like ADNI ($51.31$), LUNG\_PET\_CT\_DX ($52.86$) display lower performance. The standard deviations accompanying these scores provide insight into the consistency of the model's performance, indicating relatively small variations across individual runs. However, it's essential to note that while the model maintains consistency, there remains variability in its efficacy across different datasets, as evident in the significant deviation in scores, especially for datasets like LUNG\_PET\_CT\_DX, NLST, LIDC-IDRI, and CERA. This could be due to a number of factors, such as the quality of the data dictionary, and the specific FHIR profiles used. An informative and comprehensive dictionary could equip the LLM to interpret even cryptic column names effectively, facilitating better mapping and potentially mitigating performance variability. The variability in scores, highlights the importance of understanding dataset-specific characteristics that may influence the model's ability to accurately predict FHIR Mappings. 

Further analysis is needed to understand the factors that contribute to the model's performance on different datasets. This could involve examining the characteristics of the datasets, such as the type of data, and the quality of the data dictionary . 

\section{Future Work}\label{sec:future_work}
For our future work we have identified a few research areas:
\begin{itemize}
\item Enriching the content for context retrieval, using FHIR examples, LLM generated content etc.
\item Find out reason for lower score for certain datasets and improve the score for different datasets to get a more uniform score.
\item Work with different LLMs to have a comparative analysis of performance.
\item Work with different RAG methodologies (HyDE ~\citep{gao2022precise}, Reranking etc.) to have a more in-depth understanding of how they are impacting the score.
\end{itemize}

\section{Conclusion}\label{sec:conclusion}

    In this work, we have explored a methodology of producing FHIR mapping utilizing the inherent parameterized knowledge available in LLMs, further bolstered by external non-parameterized knowledge sourced from curated documentation of FHIR Standards. Utilizing RAG with FAISS, relevant context information is retrieved and passed to the LLM during the inference process. The results indicate that this approach is a viable method for producing FHIR mapping for data elements from clinical datasets.
    
    The distinct advantages of employing LLMs for FHIR mapping, in comparison to manual approaches, are as follows:
    
    \begin{enumerate}
        \item \textbf{Time Efficiency:} LLMs can process and map data at a significantly faster rate than manual methods, leading to time savings in the mapping process. As an example, if it takes an expert 1 minute to map a single data field to FHIR, it would take 160 hours to map the total ~9600 fields that we have mapped in the results. In our approach using LLM, we have been able to create that mapping in under 10 minutes. Usually for an expert, it will take more than a minute, considering time required for cross-referencing definition of a field, or further research for picking one resource or element over others. This is a significant advantage.
        \item \textbf{Cost of Labor:} The automation capabilities of LLMs reduce the need for extensive labor, thereby cutting down the costs associated with manual mapping. We explored Amazon Mechanical Turk pricing, which varies between \$0.02 - \$0.08 per object (in this case, we can assume an object is a single field and its FHIR mapping), at that rate our collection of 9600 fields would have cost \$192 - \$768 \citep{aws_groundtruth_pricing}. This is significantly higher than using OpenAI GPT \citep{azure_openai_pricing} and Embedding Models.
        \item \textbf{Expertise:} LLMs can effectively utilize and apply complex rules and standards, such as those in FHIR, reducing the dependency on specialized expertise. FHIR standard also gets updated with new releases, that adds an additional burden for human experts to keep themselves updated. Our automated process alleviates the need of that process.
        \item \textbf{Scalability:} LLMs offer a scalable solution that can adapt to varying sizes and complexities of clinical datasets, a task that can be challenging and resource-intensive with manual methods.
    \end{enumerate}

\section{Data Privacy and Safety considerations}
As mentioned earlier,  in course of this work, we have used only the clinical data column names and corresponding data dictionaries (which defines the columns, their data types, applicable code values etc.) where available, we have refrained from using actual clinical data as that is not a requirement for the experiments.

\section{Acknowledgement}
We would like to thank all the study participants and their families, and all the site investigators, study coordinators, and staff. Assistance in preparing this article for publication was provided by Genentech, Inc.

The authors acknowledge the National Cancer Institute and the Foundation for the National Institutes of Health, and their critical role in the creation of the free, publicly available LIDC/IDRI Database used in this study.

Alzheimer's Disease Neuroimaging Initiative (ADNI) is funded by the National Institute on Aging, the National Institute of Biomedical Imaging and Bioengineering, and through generous contributions from the following organizations: AbbVie; Alzheimer's Association; Alzheimer's Drug Discovery Foundation; Araclon Biotech; BioClinica, Inc.; Biogen; Bristol-Myers Squibb Company; CereSpir, Inc.; Cogstate; Eisai Inc.; Elan Pharmaceuticals, Inc.; Eli Lilly and Company; EuroImmun; F. Hoffmann-La Roche Ltd and its affiliated company Genentech, Inc.; Fujirebio; GE Healthcare; IXICO Ltd.; Janssen Alzheimer Immunotherapy Research \& Development, LLC.; Johnson \& Johnson Pharmaceutical Research \& Development LLC.; Lumosity; Lundbeck; Merck \& Co., Inc.; Meso Scale Diagnostics, LLC; NeuroRx Research; Neurotrack Technologies; Novartis Pharmaceuticals Corporation; Pfizer Inc.; Piramal Imaging; Servier; Takeda Pharmaceutical Company; and Transition Therapeutics. The Canadian Institutes of Health Research is providing funds to support ADNI clinical sites in Canada. Private sector contributions are facilitated by the Foundation for the National Institutes of Health (www.fnih.org). The grantee organization is the Northern California Institute for Research and Education, and the study is coordinated by the Alzheimer's Therapeutic Research Institute at the University of Southern California. ADNI data are disseminated by the Laboratory for NeuroImaging at the University of Southern California.

The Australian Imaging Biomarkers and Lifestyle (AIBL) study (www.AIBL.csiro.au) is a consortium between Austin Health, CSIRO, Edith Cowan University, the Florey Institute (The University of Melbourne), and the National Aging Research Institute. Partial financial support was provided by the Alzheimer's Association (US), the Alzheimer's Drug Discovery Foundation, an anonymous foundation, the Science and Industry Endowment Fund, the Dementia Collaborative Research Centres, the Victorian Government's Operational Infrastructure Support program, the McCusker Alzheimer's Research Foundation, the National Health and Medical Research Council, and the Yulgilbar Foundation. Numerous commercial interactions have supported data collection and analysis. In-kind support has also been provided by Sir Charles Gairdner Hospital, Cogstate Ltd., Hollywood Private Hospital, the University of Melbourne, and St. Vincent's Hospital.

The BioFINDER study was supported by the European Research Council, the Swedish Research Council, the Strategic Research Programme in Neuroscience at Lund University (MultiPark), the Crafoord Foundation, the Swedish Brain Foundation, The Swedish Alzheimer foundation, the Torsten Söderberg Foundation at the Royal Swedish Academy of Sciences, and the regional agreement on medical training and clinical research (ALF) between Region Skåne and Lund University.

The Harvard Aging Brain Study (HABS - P01AG036694; https://habs.mgh.harvard.edu) was launched in 2010, funded by the National Institute on Aging. and is led by principal investigators Reisa A. Sperling MD and Keith A. Johnson MD at Massachusetts General Hospital/Harvard Medical School in Boston, MA. 

\bibliographystyle{unsrtnat}
\bibliography{main}

\begin{thebibliography}{26}
\providecommand{\natexlab}[1]{#1}
\providecommand{\url}[1]{\texttt{#1}}
\expandafter\ifx\csname urlstyle\endcsname\relax
  \providecommand{\doi}[1]{doi: #1}\else
  \providecommand{\doi}{doi: \begingroup \urlstyle{rm}\Url}\fi

\bibitem[Modaresnezhad et~al.(2019)Modaresnezhad, Vahdati, Nemati, Ardestani,
  and Sadri]{modaresnezhad2019}
Minoo Modaresnezhad, Ali Vahdati, Hamid Nemati, Ali Ardestani, and Fereidoon
  Sadri.
\newblock {A rule-based semantic approach for data integration, standardization
  and dimensionality reduction utilizing the UMLS: Application to predicting
  bariatric surgery outcomes}.
\newblock \emph{Computers in Biology and Medicine}, 106:\penalty0 84--90, 2019.
\newblock ISSN 0010-4825.
\newblock \doi{10.1016/j.compbiomed.2019.01.019}.

\bibitem[Papadopoulos et~al.(2022)Papadopoulos, Soflano, Chaudy, Adejo, and
  Connolly]{papadopoulos2022}
Petros Papadopoulos, Mario Soflano, Yaelle Chaudy, Wilson Adejo, and Thomas~M.
  Connolly.
\newblock {A systematic review of technologies and standards used in the
  development of rule-based clinical decision support systems}.
\newblock \emph{Health and Technology}, 12\penalty0 (4):\penalty0 713--727,
  2022.
\newblock ISSN 2190-7188.
\newblock \doi{10.1007/s12553-022-00672-9}.

\bibitem[Sezgin et~al.(2023)Sezgin, Hussain, Rust, and Huang]{sezgin2023}
Emre Sezgin, Syed-Amad Hussain, Steve Rust, and Yungui Huang.
\newblock {Extracting Medical Information From Free-Text and Unstructured
  Patient-Generated Health Data Using Natural Language Processing Methods:
  Feasibility Study With Real-world Data}.
\newblock \emph{JMIR Formative Research}, 7:\penalty0 e43014, 2023.
\newblock \doi{10.2196/43014}.

\bibitem[Adamson et~al.(2023)Adamson, Waskom, Blarre, Kelly, Krismer, Nemeth,
  Gippetti, Ritten, Harrison, Ho, Linzmayer, Bansal, Wilkinson, Amster, Estola,
  Benedum, Fidyk, Estévez, Shapiro, and Cohen]{adamson2023}
Blythe Adamson, Michael Waskom, Auriane Blarre, Jonathan Kelly, Konstantin
  Krismer, Sheila Nemeth, James Gippetti, John Ritten, Katherine Harrison,
  George Ho, Robin Linzmayer, Tarun Bansal, Samuel Wilkinson, Guy Amster, Evan
  Estola, Corey~M. Benedum, Erin Fidyk, Melissa Estévez, Will Shapiro, and
  Aaron~B. Cohen.
\newblock {Approach to machine learning for extraction of real-world data
  variables from electronic health records}.
\newblock \emph{Frontiers in Pharmacology}, 14:\penalty0 1180962, 2023.
\newblock ISSN 1663-9812.
\newblock \doi{10.3389/fphar.2023.1180962}.

\bibitem[Cardinal(2017)]{cardinal2017}
Rudolf~N. Cardinal.
\newblock {Clinical records anonymisation and text extraction (CRATE): an
  open-source software system}.
\newblock \emph{BMC Medical Informatics and Decision Making}, 17\penalty0
  (1):\penalty0 50, 2017.
\newblock \doi{10.1186/s12911-017-0437-1}.

\bibitem[{Health Level Seven International}(2024)]{hl7_fhir}
{Health Level Seven International}.
\newblock Hl7 fhir.
\newblock \url{https://www.hl7.org/fhir/}, 2024.

\bibitem[{Clinical Data Interchange Standards Consortium
  (CDISC)}({2024})]{CDISC_Standards}
{Clinical Data Interchange Standards Consortium (CDISC)}.
\newblock {Standards Overview}, {2024}.

\bibitem[Petersen et~al.(2010)Petersen, Aisen, Beckett, Donohue, Gamst, Harvey,
  Jack, Jagust, Shaw, Toga, et~al.]{petersen2010alzheimer}
Ronald~Carl Petersen, PS~Aisen, Laurel~A Beckett, MC~Donohue, AC~Gamst,
  Danielle~J Harvey, CR~Jack, WJ~Jagust, LM~Shaw, AW~Toga, et~al.
\newblock Alzheimer's disease neuroimaging initiative (adni): clinical
  characterization.
\newblock \emph{Neurology}, 74\penalty0 (3):\penalty0 201--209, 2010.

\bibitem[{A4 Study team}(Accessed: 2024)]{a4study}
{A4 Study team}.
\newblock {A4 Study Data}.
\newblock \url{https://a4study.org/}, Accessed: 2024.
\newblock Date accessed: [24 January 2024].

\bibitem[Ellis et~al.(2009)Ellis, Nathan, Villemagne, Mulligan, Saunder, Young,
  Smith, Welch, Woodward, Wesnes, et~al.]{ellis2009galantamine}
JR~Ellis, Pradeep~Jonathan Nathan, Victor~L Villemagne, RS~Mulligan, Timothy
  Saunder, K~Young, Catherine~L Smith, J~Welch, Michael Woodward, Keith~A
  Wesnes, et~al.
\newblock Galantamine-induced improvements in cognitive function are not
  related to alterations in $\alpha$4$\beta$2 nicotinic receptors in early
  alzheimer’s disease as measured in vivo by 2-[18f] fluoro-a-85380 pet.
\newblock \emph{Psychopharmacology}, 202\penalty0 (1):\penalty0 79--91, 2009.

\bibitem[adn(Accessed: 2024)]{adnidod}
{ADNI DOD Study}.
\newblock \url{https://ncrad.iu.edu/resource/adni_dod.html}, Accessed: 2024.
\newblock Date accessed: [24 January 2024].

\bibitem[Mattsson-Carlgren et~al.(2020)Mattsson-Carlgren, Leuzy, Janelidze,
  Palmqvist, Stomrud, Strandberg, Smith, and Hansson]{mattsson2020implications}
Niklas Mattsson-Carlgren, Antoine Leuzy, Shorena Janelidze, Sebastian
  Palmqvist, Erik Stomrud, Olof Strandberg, Ruben Smith, and Oskar Hansson.
\newblock The implications of different approaches to define at (n) in
  alzheimer disease.
\newblock \emph{Neurology}, 94\penalty0 (21):\penalty0 e2233--e2244, 2020.

\bibitem[Teng et~al.(2022)Teng, Manser, Pickthorn, Brunstein, Blendstrup,
  Sanabria~Bohorquez, Wildsmith, Toth, Dolton, Ramakrishnan, Bobbala, Sikkes,
  Ward, Fuji, Kerchner, and Investigators]{teng2022safety}
E.~Teng, P.~T. Manser, K.~Pickthorn, F.~Brunstein, M.~Blendstrup,
  S.~Sanabria~Bohorquez, K.~R. Wildsmith, B.~Toth, M.~Dolton, V.~Ramakrishnan,
  A.~Bobbala, S.~A.~M. Sikkes, M.~Ward, R.~N. Fuji, G.~A. Kerchner, and Tauriel
  Investigators.
\newblock Safety and efficacy of semorinemab in individuals with prodromal to
  mild alzheimer disease: A randomized clinical trial.
\newblock \emph{JAMA Neurol}, 79\penalty0 (8):\penalty0 758--767, 2022.
\newblock \doi{10.1001/jamaneurol.2022.1375}.
\newblock PMID: 35696185; PMCID: PMC9194753.

\bibitem[Dagley et~al.(2017)Dagley, LaPoint, Huijbers, Hedden, McLaren,
  Chatwal, Papp, Amariglio, Blacker, Rentz, et~al.]{dagley2017harvard}
Alexander Dagley, Molly LaPoint, Willem Huijbers, Trey Hedden, Donald~G
  McLaren, Jasmeer~P Chatwal, Kathryn~V Papp, Rebecca~E Amariglio, Deborah
  Blacker, Dorene~M Rentz, et~al.
\newblock Harvard aging brain study: dataset and accessibility.
\newblock \emph{Neuroimage}, 144:\penalty0 255--258, 2017.

\bibitem[Li et~al.(2020)Li, Wang, Li, Lu, HuangFu, and Wang]{li2020large}
P.~Li, S.~Wang, T.~Li, J.~Lu, Y.~HuangFu, and D.~Wang.
\newblock A large-scale ct and pet/ct dataset for lung cancer diagnosis
  (lung-pet-ct-dx).
\newblock \url{https://doi.org/10.7937/TCIA.2020.NNC2-0461}, 2020.
\newblock Date accessed: Enter date here.

\bibitem[Clark et~al.(2013)Clark, Vendt, Smith, Freymann, Kirby, Koppel, Moore,
  Phillips, Maffitt, Pringle, Tarbox, and Prior]{clark2013cancer}
K.~Clark, B.~Vendt, K.~Smith, J.~Freymann, J.~Kirby, P.~Koppel, S.~Moore,
  S.~Phillips, D.~Maffitt, M.~Pringle, L.~Tarbox, and F.~Prior.
\newblock The cancer imaging archive (tcia): Maintaining and operating a public
  information repository.
\newblock \emph{Journal of Digital Imaging}, 26\penalty0 (6):\penalty0
  1045--1057, 2013.
\newblock \doi{10.1007/s10278-013-9622-7}.

\bibitem[Armato et~al.(2015)Armato, McLennan, Bidaut, McNitt-Gray, Meyer,
  Reeves, Zhao, Aberle, Henschke, Hoffman, Kazerooni, MacMahon, Van~Beek,
  Yankelevitz, Biancardi, Bland, Brown, Engelmann, Laderach, Max, Pais, Qing,
  Roberts, Smith, Starkey, Batra, Caligiuri, Farooqi, Gladish, Jude, Munden,
  Petkovska, Quint, Schwartz, Sundaram, Dodd, Fenimore, Gur, Petrick, Freymann,
  Kirby, Hughes, Casteele, Gupte, Sallam, Heath, Kuhn, Dharaiya, Burns, Fryd,
  Salganicoff, Anand, Shreter, Vastagh, Croft, and Clarke]{armato2015data}
S.~G.~III Armato, G.~McLennan, L.~Bidaut, M.~F. McNitt-Gray, C.~R. Meyer, A.~P.
  Reeves, B.~Zhao, D.~R. Aberle, C.~I. Henschke, E.~A. Hoffman, E.~A.
  Kazerooni, H.~MacMahon, E.~J.~R. Van~Beek, D.~Yankelevitz, A.~M. Biancardi,
  P.~H. Bland, M.~S. Brown, R.~M. Engelmann, G.~E. Laderach, D.~Max, R.~C.
  Pais, D.~P.~Y. Qing, R.~Y. Roberts, A.~R. Smith, A.~Starkey, P.~Batra,
  P.~Caligiuri, A.~Farooqi, G.~W. Gladish, C.~M. Jude, R.~F. Munden,
  I.~Petkovska, L.~E. Quint, L.~H. Schwartz, B.~Sundaram, L.~E. Dodd,
  C.~Fenimore, D.~Gur, N.~Petrick, J.~Freymann, J.~Kirby, B.~Hughes, A.~V.
  Casteele, S.~Gupte, M.~Sallam, M.~D. Heath, M.~H. Kuhn, E.~Dharaiya,
  R.~Burns, D.~S. Fryd, M.~Salganicoff, V.~Anand, U.~Shreter, S.~Vastagh, B.~Y.
  Croft, and L.~P. Clarke.
\newblock Data from lidc-idri, 2015.
\newblock URL \url{https://doi.org/10.7937/K9/TCIA.2015.LO9QL9SX}.
\newblock Date accessed: [Add access date here].

\bibitem[Armato et~al.(2011)Armato, McLennan, Bidaut, McNitt-Gray, Meyer,
  Reeves, Zhao, Aberle, Henschke, Hoffman, Kazerooni, MacMahon, Van~Beeke,
  Yankelevitz, Biancardi, Bland, Brown, Engelmann, Laderach, Max, Pais, Qing,
  Roberts, Smith, Starkey, Batrah, Caligiuri, Farooqi, Gladish, Jude, Munden,
  Petkovska, Quint, Schwartz, Sundaram, Dodd, Fenimore, Gur, Petrick, Freymann,
  Kirby, Hughes, Casteele, Gupte, Sallamm, Heath, Kuhn, Dharaiya, Burns, Fryd,
  Salganicoff, Anand, Shreter, Vastagh, and Croft]{armato2011lung}
S.~G.~3rd Armato, G.~McLennan, L.~Bidaut, M.~F. McNitt-Gray, C.~R. Meyer, A.~P.
  Reeves, B.~Zhao, D.~R. Aberle, C.~I. Henschke, E.~A. Hoffman, E.~A.
  Kazerooni, H.~MacMahon, E.~J. Van~Beeke, D.~Yankelevitz, A.~M. Biancardi,
  P.~H. Bland, M.~S. Brown, R.~M. Engelmann, G.~E. Laderach, D.~Max, R.~C.
  Pais, D.~P. Qing, R.~Y. Roberts, A.~R. Smith, A.~Starkey, P.~Batrah,
  P.~Caligiuri, A.~Farooqi, G.~W. Gladish, C.~M. Jude, R.~F. Munden,
  I.~Petkovska, L.~E. Quint, L.~H. Schwartz, B.~Sundaram, L.~E. Dodd,
  C.~Fenimore, D.~Gur, N.~Petrick, J.~Freymann, J.~Kirby, B.~Hughes, A.~V.
  Casteele, S.~Gupte, M.~Sallamm, M.~D. Heath, M.~H. Kuhn, E.~Dharaiya,
  R.~Burns, D.~S. Fryd, M.~Salganicoff, V.~Anand, U.~Shreter, S.~Vastagh, and
  B.~Y. Croft.
\newblock The lung image database consortium (lidc) and image database resource
  initiative (idri): A completed reference database of lung nodules on ct
  scans.
\newblock \emph{Medical Physics}, 38:\penalty0 915--931, 2011.
\newblock \doi{https://doi.org/10.1118/1.3528204}.

\bibitem[Guymer et~al.(2019)Guymer, Wu, Hodgson, Caruso, Brassington, Tindill,
  Aung, McGuinness, Fletcher, Chen, Chakravarthy, Arnold, Heriot, Durkin, Lek,
  Harper, Wickremasinghe, Sandhu, Baglin, Sharangan, Braat, and Luu]{cera}
Robyn~H. Guymer, Zhichao Wu, Lauren~A.B. Hodgson, Emily Caruso, Kate~H.
  Brassington, Nicole Tindill, Khin~Zaw Aung, Myra~B. McGuinness, Erica~L.
  Fletcher, Fred~K. Chen, Usha Chakravarthy, Jennifer~J. Arnold, Wilson~J.
  Heriot, Shane~R. Durkin, Jia~Jia Lek, Colin~A. Harper, Sanjeewa~S.
  Wickremasinghe, Sukhpal~S. Sandhu, Elizabeth~K. Baglin, Pyrawy Sharangan,
  Sabine Braat, and Chi~D. Luu.
\newblock Subthreshold nanosecond laser intervention in age-related macular
  degeneration: The lead randomized controlled clinical trial, June 2019.

\bibitem[{Citeline}(2024)]{CitelineRWD}
{Citeline}.
\newblock {Real World Data Solutions}, 2024.
\newblock URL
  \url{https://www.citeline.com/en/solutions/clinical/real-world-data-solutions}.

\bibitem[Lewis et~al.(2020)Lewis, Perez, Piktus, Petroni, Karpukhin, Goyal,
  K{\"{u}}ttler, Lewis, Yih, Rockt{\"{a}}schel, Riedel, and
  Kiela]{DBLP:journals/corr/abs-2005-11401}
Patrick S.~H. Lewis, Ethan Perez, Aleksandra Piktus, Fabio Petroni, Vladimir
  Karpukhin, Naman Goyal, Heinrich K{\"{u}}ttler, Mike Lewis, Wen{-}tau Yih,
  Tim Rockt{\"{a}}schel, Sebastian Riedel, and Douwe Kiela.
\newblock Retrieval-augmented generation for knowledge-intensive {NLP} tasks.
\newblock \emph{CoRR}, abs/2005.11401, 2020.

\bibitem[{HL7 FHIR Resource List}(2023)]{FHIRv5Resources}
{HL7 FHIR Resource List}, 2023.
\newblock URL \url{https://www.hl7.org/fhir/resourcelist.html}.

\bibitem[Douze et~al.(2024)Douze, Guzhva, Deng, Johnson, Szilvasy, Mazaré,
  Lomeli, Hosseini, and Jégou]{faiss}
Matthijs Douze, Alexandr Guzhva, Chengqi Deng, Jeff Johnson, Gergely Szilvasy,
  Pierre-Emmanuel Mazaré, Maria Lomeli, Lucas Hosseini, and Hervé Jégou.
\newblock The faiss library.
\newblock 2024.

\bibitem[Gao et~al.(2022)Gao, Ma, Lin, and Callan]{gao2022precise}
Luyu Gao, Xueguang Ma, Jimmy Lin, and Jamie Callan.
\newblock Precise zero-shot dense retrieval without relevance labels, 2022.

\bibitem[{AWS}(2023)]{aws_groundtruth_pricing}
{AWS}.
\newblock Amazon sagemaker ground truth – pricing.
\newblock \url{https://aws.amazon.com/sagemaker/groundtruth/pricing/}, 2023.

\bibitem[{Azure OpenAI}(2023)]{azure_openai_pricing}
{Azure OpenAI}.
\newblock Azure cognitive services openai service pricing.
\newblock
  \url{https://azure.microsoft.com/en-us/pricing/details/cognitive-services/openai-service/},
  2023.

\end{thebibliography}

\end{document}